\documentclass[final,3p,times,twocolumn,authoryear]{elsarticle}

\usepackage{amssymb}
\usepackage{amsmath}
\usepackage{float}
\usepackage{hyperref}
\usepackage{url}
\usepackage{lineno}
\usepackage{multirow}%
\providecommand{\URLprefix}{URL: }
\providecommand{\DOIprefix}{doi: }

\providecommand{\doi}[1]{\href{https://doi.org/#1}{#1}}
\usepackage{amssymb}
\usepackage{booktabs}
\usepackage{subcaption}

\journal{Waste Management Journal}

\begin{document}

\begin{frontmatter}



\title{WasteAssistant: Regulation-Guided Visual Question Answering Framework for Intelligent Waste Segregation and Sustainable Management} 
\author[inst1]{Khush Kataruka}
\ead{u23ai112@coed.svnit.ac.in}
\author[inst1]{Harshit Maurya}
\ead{u23ai122@coed.svnit.ac.in}
\author[inst2]{Anuja Vats}
\ead{anuja.vats@ntnu.no}
\author[inst3]{Murari Mandal}
\ead{murari.mandalfcs@kiit.ac.in}
\author[inst2]{Kiran Raja}
\ead{kiran.raja@ntnu.no}
\author[inst1]{Praveen Kumar Chandaliya\corref{cor1}}
\ead{pkc@ai.svnit.ac.in}
\cortext[cor1]{Corresponding author}
\affiliation[inst1]{
	organization={Sardar Vallabhbhai National Institute of Technology},
	city={Surat},
	postcode={395007},
	state={Gujarat},
	country={India}
}
\affiliation[inst2]{
	organization={Norwegian University of Science and Technology},
	city={Gjøvik},
	country={Norway}
}
\affiliation[inst3]{
	organization={Kalinga Institute of Industrial Technology},
	city={Bhubaneswar},
	state={Odisha},
	country={India}
}

\fntext[fn1]{Source code and WasteVQA dataset are publicly available at: 
	\url{https://github.com/Khushkataruka/WasteAssistant}}
\begin{abstract}
Efficient waste segregation is critical for sustainable urban management and environmental governance. Existing automated systems are limited by single-modality visual processing, insufficient contextual understanding, and weak regulatory alignment. To address these issues, we propose a language-guided vision-AI framework that integrates vision-language models and multimodal large language models for joint visual-linguistic reasoning. This framework implements a visual question answering paradigm aligned with India’s Solid Waste Management Rules 2016. We construct a new WasteVQA dataset with 13,500 question-answer pairs across 21 waste categories. Experiments show that the BLIP-based model achieves a BLEU score of 0.8291 and a BERTScore of 0.9273, outperforming traditional CNN-based methods. This work improves source-level segregation accuracy, ensures regulatory compliance, and supports scalable deployment for municipal and citizen-facing waste management, promoting multimodal AI in sustainable urban infrastructure. The source code and dataset are available at:
\url{https://github.com/Khushkataruka/WasteAssistant}
\end{abstract}

%



\end{frontmatter}

\section{Introduction}
Effective solid waste management has become a critical global challenge, with the world generating over $2$ billion tons of municipal solid waste annually~\citep{worldbank_waste}. This figure is projected to increase by $70\%$ to $3.8$ billion tons by $2050$~\citep{worldbank_waste,un_swm_outlook}. This escalating waste crisis poses a significant threat to the environment, public health, and economic development. Improperly managed waste contributes to air, water, and soil pollution and is a major source of methane, a potent greenhouse gas. Furthermore, it can lead to the spread of diseases and respiratory problems, particularly in low-income countries where over $90\%$ of waste is often disposed of in unregulated dumps or openly burned. The economic consequences are also severe, with the cost of inaction projected to exceed USD $600$ billion per year by $2050$ \citep{worldbank_waste}. A major contributor to this crisis is inefficient waste classification, which undermines recycling efforts, accelerates environmental degradation, and endangers both wildlife and human communities~\citep{cheng2023impact}. Animals frequently ingest or come into contact with hazardous materials due to misclassified waste, leading to life-threatening health effects. Moreover, manual sorting methods are labour-intensive, error-prone, and unable to scale with growing urban demands~\citep{gundupalli2017review}.

To address the pressing challenges of modern waste management, automated waste classification systems leveraging deep learning techniques~\citep{JIN2023123} and IoT-enabled smart sensing infrastructures are increasingly adopted for real-time, scalable waste monitoring~\citep{WANG202120}. Driven by recent advancements in Vision Language Model, these systems offer scalable, accurate, and efficient solutions that enhance waste segregation, improve recycling rates, and reduce ecological and health risks. In this paper, we propose a novel approach based on Visual Question Answering (VQA) to support waste segregation in accordance with India’s Solid Waste Management (SWM) Rules, 2016~\citep{swm2016}. This alignment ensures standardized waste categorization and facilitates practical, policy-compliant deployment in real-world settings. Our primary contribution is the development of a VQA model fine-tuned on ``WasteVQA'' a newly curated, regulation-compliant dataset designed to generate actionable guidance for responsible waste disposal. Unlike conventional classification models that only categorise waste types, our system not only identifies waste items from visual input but also provides nuanced responses to natural language queries about their appropriate handling and disposal.

By integrating Vision-Language Learning Models (VLLMs) with a domain-specific dataset, this approach bridges the gap between image recognition and regulatory reasoning, enabling an intelligent, interpretable, and context-aware solution for waste management. To address these challenges, automated waste classification systems guided by advancements in vision and language-based AI are becoming essential. These systems enhance disposal efficiency, improve recycling rates, and reduce ecological risks through accurate and scalable segregation.

Our main contribution is a VQA model trained on ``WasteVQA'' a novel, regulation-specific dataset designed to provide actionable guidance on waste disposal. We demonstrate that by fine-tuning a Vision-Language Model (VLLM) on this specialized dataset, we can create a powerful tool that not only identifies waste but also provides crucial information on its proper handling and disposal, thereby improving upon the limitations of standard classification. The primary objectives of this paper are:
\begin{enumerate}
	\item We presented the WasteVQA dataset, the first regulation aware VQA dataset for solid waste management, structured around India’s Solid Waste Management (SWM) Rules,2016, enabling standardized and policy-compliant waste categorization. The dataset comprises $13,500$ question-answer pairs across $21$ waste categories, including challenging types such as sanitary, hazardous, and e-waste.
	\item We perform task-specific fine-tuning and comprehensive evaluation of state-of-the-art VQA models using the proposed dataset, examining their effectiveness in interpreting visual waste content and producing accurate, regulation-compliant disposal recommendations.
	\item The framework is evaluated against state-of-the-art VQA baselines such as BLIP and InstructBLIP under uniform training settings, demonstrating its effectiveness in regulatory-aligned question answering.
	\item The proposed system is designed to support real-world deployment by municipalities and stakeholders, facilitating explainable, scalable, and policy-compliant AI-assisted waste segregation and monitoring.
\end{enumerate}

\section{Related Work}
In this section, we review the evolution of automated solid waste management research along two primary directions: (i) deep learning–based approaches, which concentrate on visual recognition of waste categories using Convolutional Neural Networks (CNNs) and their variants, and (ii) multi-modal large language model-  (MLLM) approaches, which extend beyond visual classification to support reasoning over both images and textual queries.
\subsection{\textbf{Deep Learning}}
Automated waste classification has emerged as a critical research area, fueled by the urgent need for scalable and efficient waste management solutions. The foundational efforts in this field initially relied on traditional machine learning algorithms such as Support Vector Machines (SVM) \citep{Uganya2022} and Random Forests, which necessitated handcrafted features and demonstrated limited adaptability in diverse, real-world environments \citep{yang2016waste,ml_waste_review}. These early models were constrained in terms of scalability and robustness, making them unsuitable for the complexity of modern waste streams.

The breakthrough came with the application of CNNs, which enabled the automatic extraction of features from raw images. The introduction of the TrashNet dataset by Yang et al.~\citep{yang2016waste} provided a pivotal benchmark, consisting of six waste categories: paper, plastic, metal, glass, cardboard, and general trash. This spurred a wave of research exploring CNN-based models. For instance, Adedeji et. al~\citep{adedeji2020accurate} used ResNet-50 as a feature extractor followed by an SVM classifier, achieving $87\%$ accuracy. Mao et al.~\citep{mao2021integrating} proposed a hybrid approach that integrates DenseNet121 with a genetic algorithm, improving the accuracy to $94.02\%$. Vo et al.~\citep{vo2021deep} presented DNN-TC, based on an improved ResNeXt model, reaching $94\%$ classification accuracy. While these models reported strong results, their reliance on static datasets, such as TrashNet, limited their applicability. Real-world waste images often include occlusions, multiple objects, poor lighting, or deformed items, challenges not well-represented in benchmark datasets. To bridge this gap, Chu et al.~\citep{chu2021multilayer} proposed a hybrid architecture that combines CNNs with multi-layer perceptrons and integrates data from embedded sensors. Their system demonstrated over $90\%$ accuracy but introduced significant complexity and hardware dependencies. Another research trend focused on domain-specific classification. Bobulski et al.~\citep{bobulski2020classification} employed deep learning to classify different types of plastic, a task crucial for recycling. Zhou et al.~\citep{zhou2022medical} and Hossain et al.~\citep{hossain2022rwc} targeted medical waste using enhanced ResNeXt and deep ensemble networks, respectively. 
Despite these advances, CNN-based systems remain limited to fixed class labels and are unable to incorporate contextual, rule-based, or policy-driven decision-making. Although such approaches highlighted the importance of fine-grained classification, they were often restricted to narrowly defined waste categories and exhibited limited generalizability to real-world scenarios. To overcome these limitations, recent research has shifted toward MLLMs, which combine visual understanding with natural language reasoning.
\subsection{\textbf{Multi-Modal Large Language Model}}
Recent advances in MLLMs~\citep{alayrac2022flamingo,li2022blip,dai2023instructblip,liu2023llava} have expanded vision-based AI far beyond static classification. Unlike CNN-based models that predict fixed categories, MLLMs fuse visual and textual modalities, enabling instruction-following, context-sensitive reasoning, and grounded response generation. This shift has driven-progress in tasks such as captioning, visual grounding, and, most notably, VQA. First formalized by Antol et al.~\citep{antol2015vqa}, VQA introduced benchmarks for answering natural language questions about images. The field quickly evolved: surveys by Kafle and Kanan \citep{kafle2017visual} summarized datasets and challenges, while Goyal et al. \citep{goyal2016making} exposed biases that spurred models emphasizing genuine visual understanding. Subsequent advances, such as Bilinear Attention Networks \citep{kim2018bilinear}, BLIP \citep{li2022blip}, and instruction-tuned models like InstructBLIP \citep{dai2023instructblip}, LLaVA \citep{liu2023llava}, and more recently Qwen \citep{bai2023qwen}, achieved stronger image–text alignment, complex reasoning, and human-aligned language generation. VQA now serves as a benchmark for measuring multimodal perception and reasoning, with applications requiring interpretability and compliance beyond simple recognition.

Although VQA has been applied in agriculture and plant pathology \citep{jain2019agribot,bhavika2020agribot,guofeng2020question,lan2023visual,lu2024application}, its potential in waste management and environmental monitoring remains largely untapped. Agricultural VQA demonstrates how domain-specific datasets and expert Q\&A templates improve diagnostic accuracy and utility. Extending these insights to waste management could yield systems that not only classify waste but also answer compliance-related questions, for instance, determining recyclability, detecting hazards, or suggesting disposal methods. Such capabilities would directly support smart city planning, circular economy practices, and sustainability goals, underscoring the need for research into domain-specific multimodal datasets and instruction-tuned VQA models.
\begin{figure*}[h!]
	\centering
	\includegraphics[width=\textwidth]{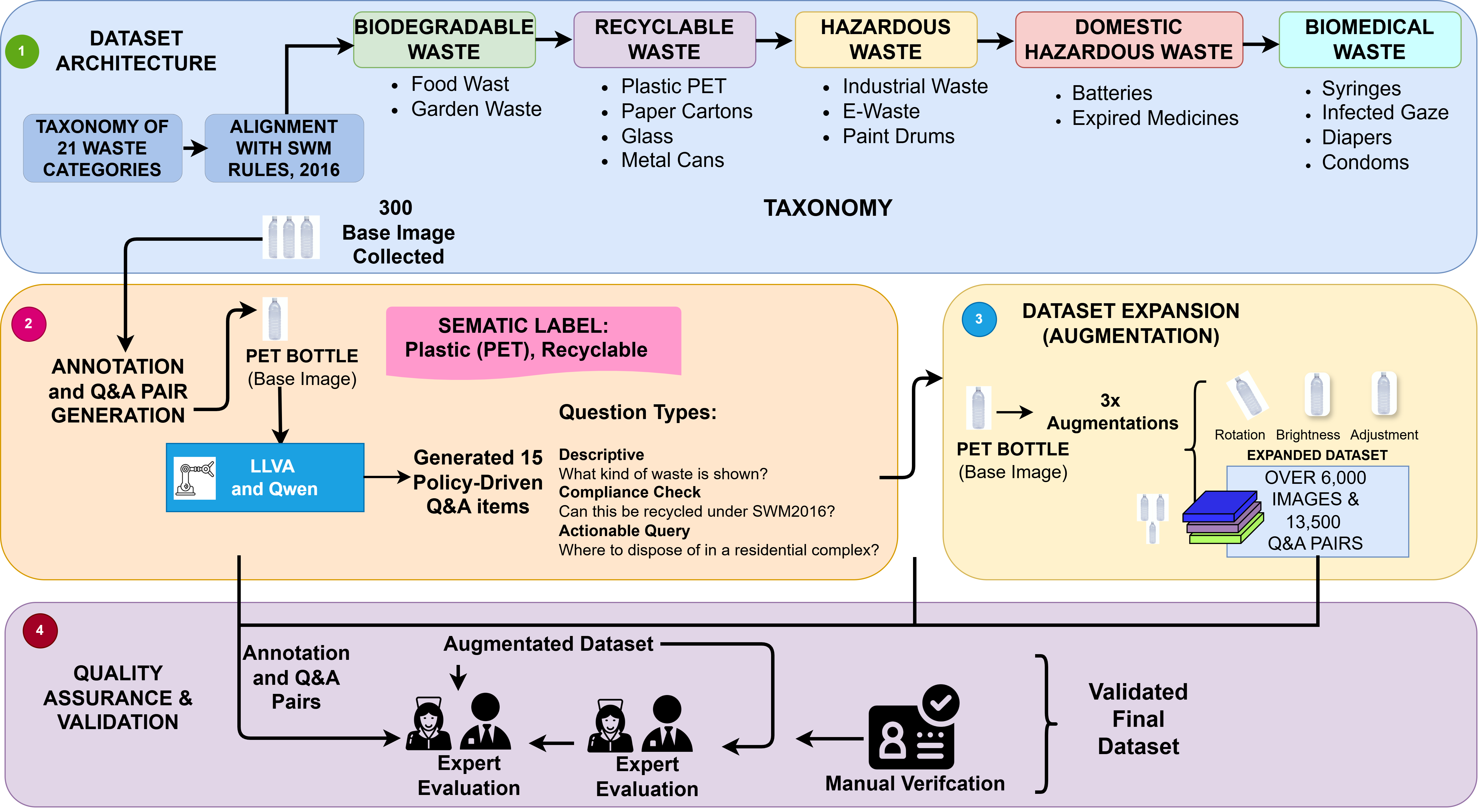}
	\caption{Overall data curation workflow}
	\label{fig:workflow}
\end{figure*}
\section{Dataset Curation}
As previously stated, WasteVQA is a domain-specific Visual Question Answering (VQA) dataset developed in alignment with India’s Solid Waste Management (SWM) Rules, 2016~\citep{swm2016}, which constitute a legally mandated and nationally recognized framework for waste categorization, handling, and disposal.
To ensure both visual diversity and comprehensive category coverage, images were aggregated from multiple publicly available repositories, including TrashNet~\citep{yang2016waste}, TrashBox~\citep{kumsetty2021trashbox}, and BDwaste~\citep{rahman2024bdwaste}. This integration enabled a broad and representative coverage of $21$ waste categories commonly encountered in domestic and municipal settings, consistent with the classifications defined under the SWM Rules. The overall dataset curation workflow Fig~\ref{fig:workflow} comprised the following key stages:
\begin{enumerate}
	\item \textbf{Category Definition:}  A taxonomy of 21 waste categories was designed in alignment with SWM Rules, 2016, covering biodegradable, recyclable, hazardous, biomedical, sanitary, and domestic hazardous waste types.
	\item \textbf{Image Collection and Annotation:} Approximately 300 base images were collected across these categories and annotated with both semantic labels and regulatory metadata.
	\item \textbf{Q\&A Pair Generation:} Each base image was paired with 15 policy-driven Q\&A items generated using LLAVA and Qwen2 models. Questions  spanned descriptive queries (e.g., “What kind of waste is shown?”), compliance checks (e.g., “Can this be recycled under SWM 2016?”), and actionable queries (e.g., “Where should this be disposed of in a residential complex?”). Questions were diversified in phrasing, complexity (yes/no, multiple choice, open-ended)
	
	\item \textbf{Data Augmentation:} To improve robustness, we applied three standard data augmentation operations, rotation and brightness adjustment, resulting in an expanded dataset comprising over $6,000$ images and $13,500$ Q\&A pairs.
	
	
	\item \textbf{Validation and Quality Assurance:}  All annotations and question–answer pairs were subjected to a two-stage validation process, involving expert evaluation by environmental engineers and Solid Waste Management (SWM) specialists, followed by manual verification to ensure factual correctness, consistency, and regulatory alignment.
	
\end{enumerate}
\textbf{WasteVQA} comprises $6,000+$ images and $13,500$  Q\&A pairs across $21$ waste categories, annotated in JSON and CSV formats. As the first domain-specific VQA dataset for solid waste management, it embeds SWM 2016 regulatory knowledge, enabling models to move beyond recognition toward actionable, policy-aware reasoning answering not only \textit{what} an object is, but also \textit{what should be done with it}.

\begin{figure*}[h!]
	\centering
	\includegraphics[width=\textwidth]{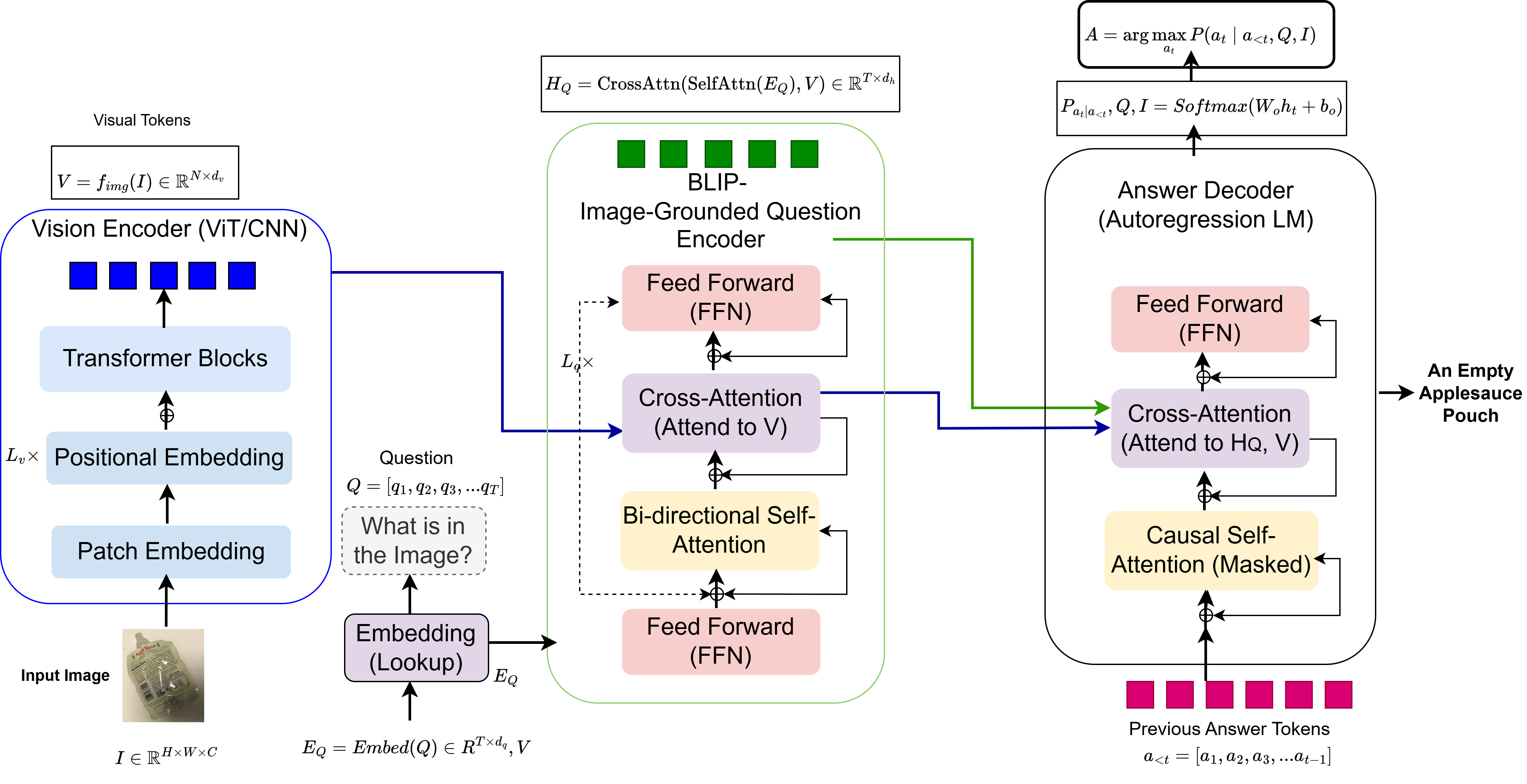}
	\caption{The core VQA model architecture, showing the Image Encoder, Image Grounded Question Encoder, and Answer Decoder.}
	\label{fig:architecture}
\end{figure*}
\section{Propose Model Architecture}
As illustrated in Fig~\ref{fig:architecture}, the presented model architecture consists of three main components: the image encoder, the image-grounded question encoder, and the answer decoder.
\subsection{\textbf{Image Encoder $(f_{\text{img}}(I))$}} Image Encoder processes visual inputs to extract meaningful feature representations from raw images using deep visual vision transformers. It converts an image $i$ ($i \sim I \in \mathbb{R}^{H \times W \times C}$) into a sequence of embeddings that capture spatial, semantic, and object-level information, providing a rich visual foundation for understanding subsequent questions and generating answers. The input image be $I \in \mathbb{R}^{H \times W \times C}$
where $H$, $W$, and $C$ are the height, width, and number of channels. The image encoder $f_{\text{img}}$ extracts visual features: $ V = f_{\text{img}}(I) \in \mathbb{R}^{N \times d_v},
$ where $ V = [v_1, v_2, \dots, v_N] $ is a sequence of $N$ visual embeddings, each of dimension $d_v$.

\subsection{\textbf{Image-Grounded Question Encoder}} Image-Grounded Question Encoder encodes the input question while incorporating the context of the image. It first applies self-attention to capture relationships between words in the question and then uses feed-forward layers to refine these representations. By aligning the question embeddings with the encoded image features, this module enables the model to interpret queries in the context of the visual content. The input question tokens be $Q = [q_1, q_2, \dots, q_T],$
where $T$ is the sequence length. The Question Encoder $f_q$ maps the tokens into an embedding, as expressed in Eq.~\ref{equ:emb}.
\begin{equation}
	\centering
	E_Q = \operatorname{Embed}(Q) \in \mathbb{R}^{T \times d_q}.
	\label{equ:emb}
\end{equation}

Next, the encoder integrates image features with question embeddings using self-attention and cross-attention, as expressed in Eq.~\ref{eq:hq}.
\begin{equation}
	\label{eq:hq}
	H_Q = \text{CrossAttention}(\text{SelfAttention}(E_Q), V) \in \mathbb{R}^{T \times d_h},
\end{equation}


where $H_Q$ are the contextualized question embeddings aligned with image features.

\subsection{Answer Decoder} Answer Decoder fuses the encoded image features with the question embeddings to generate a response. It employs bi-directional self-attention to fully understand the question, cross-attention to link relevant image regions with question tokens, and feed-forward layers to refine the fused representation. The final answer generation (Causal Decoder) then autoregressively produces the answer, generating one token at a time. It uses causal self-attention to ensure that each token depends only on previously generated tokens, cross-attention to ground the output in both the image and question features, and feed-forward layers to refine the generation process. Together, these components enable the model to reason effectively over both visual and textual inputs.
The Answer Decoder $f_{\text{dec}}$ fuses $V$ and $H_Q$ to generate the answer sequence 
$
A = [a_1, a_2, \dots, a_L]
$
of length $L$.  The causal decoder operates in an autoregressive manner, as expressed in Eqs.~\ref{eq:decoder}, \ref{eq:softmax}, and~\ref{eq:p}.

\begin{equation}
	\begin{aligned}
		h_t &= \operatorname{DecoderLayer}(h_{t-1}, H_Q, V), 
		&& t = 1,\dots,L
		\label{eq:decoder}
	\end{aligned}
\end{equation}


\begin{equation}
	P(a_t \mid a_{<t}, Q, I) = \text{Softmax}(W_o h_t + b_o)
	\label{eq:softmax}
\end{equation}
\begin{equation}
	A = \arg\max_{a_t} P(a_t \mid a_{<t}, Q, I)
	\label{eq:p}
\end{equation}
where $h_0 = \operatorname{InitialState}$, $W_o$ and $b_o$ are the output projection parameters, and $P(a_t \mid a_{<t}, Q, I)$ gives the probability of the next token conditioned on previous tokens, the question, and the image.

\textbf{Overall Objective Function} 
The full model can be compactly expressed as shown in Eq.~\ref{eq:overall}.
\begin{equation}
	A = f_{\text{model}}(I, Q) = f_{\text{dec}}\big(f_q(Q, f_{\text{img}}(I)), f_{\text{img}}(I)\big)
	\label{eq:overall}
\end{equation}
This formulation illustrates the hierarchical dependency in which the image is first encoded, the question is grounded in the visual features, and the decoder generates the final answer in an autoregressive manner.
\begin{figure*}[h!]
	\centering
	\includegraphics[width=\textwidth]{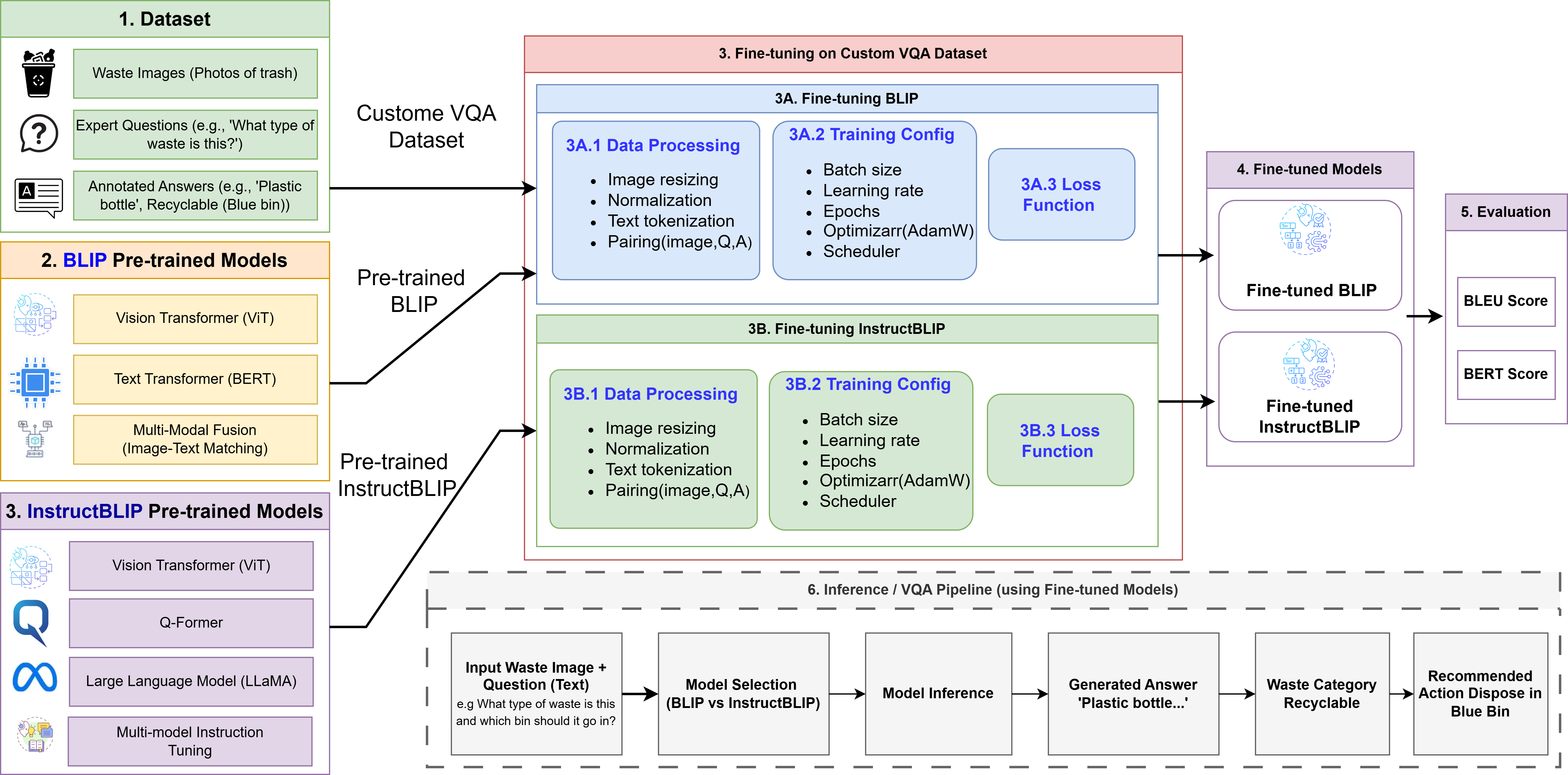}
	\caption{VQA Pipeline using fine-tuned BLIP and InstructBLIP for Weste segregation.}
	\label{fig:pipeline}
\end{figure*}
\section{VQA Pipeline for Waste Segregation}
As VQA Pipeline illustrate in Fig~\ref{fig:pipeline},
The proposed system is designed to intelligently identify waste materials and answer user questions about proper waste disposal using advanced vision-language models (VLMs) such as BLIP and InstructBLIP. The pipeline starts by preparing a custom dataset containing waste images, expert-generated questions, and corresponding annotated answers. These image-question-answer pairs are used to train the models for Visual Question Answering (VQA) tasks. In the BLIP framework, a Vision Transformer (ViT) extracts visual information from waste images, a text transformer processes the user’s question, and the two are combined through multimodal fusion to generate meaningful answers. InstructBLIP further enhances this process by integrating a Q-Former and a Large Language Model (LLaMA), enabling better reasoning and instruction-following capabilities. Both models are then fine-tuned on the custom waste-management dataset using preprocessing techniques such as image resizing, normalization, and text tokenization, along with optimized training settings including learning rate, batch size, optimizer, scheduler, and loss functions. After training, the fine-tuned models are evaluated using BLEU and BERT scores to assess the quality and semantic accuracy of the generated responses. During real-time inference, a user uploads a waste image and asks a question such as “What type of waste is this and which bin should it go into?” The selected model analyzes the image and question, predicts the waste category, generates an appropriate response, and recommends the correct disposal action, such as placing the item in the blue recycling bin. This framework can support smart waste segregation, recycling automation, and sustainable waste management applications in smart cities and environmental monitoring systems.

\section{Experiments Setup and Results}
This section presents the model training details and evaluation results used to validate the proposed model's effectiveness.
\subsection{\textbf{Model Training Details}}
In this study, we employed the BLIP and InstructBLIP models and adopted a two-stage fine-tuning strategy to enable effective domain adaptation.  Specifically, BLIP and InstructBLIP were selected because they are widely adopted vision language foundation models that represent strong and well-established baselines for image–text understanding and are suitable for question-answering and multimodal reasoning scenarios. Their inclusion enables a fair and meaningful comparison with our approach.  In the first stage, the vision encoder was frozen for the initial $10$ epochs, allowing the text and fusion modules to adapt to our specialized question–answer (Q\&A) format while preserving the pre-trained visual representations. In the second stage, the vision encoder was unfrozen for the remaining epochs, enabling full end-to-end fine-tuning and allowing the model to learn the visual characteristics specific to our waste dataset. Training was conducted using the following hyperparameters AdamW optimizer, $50$ total epochs, a learning rate of $4 \times 10^{-5}$, ExponentialLR scheduler, weight decay of $5 \times 10^{-6}$, and a batch size of $32$. These hyperparameter settings enabled stable convergence and effective domain-specific adaptation. In Fig~\ref{fig:blip} and Fig~\ref{fig:instructblip_grid}.
\begin{table*}[h!]
	\centering
	\caption{Comparison of Vision-Language Model Architectures Used in WasteAssistant}
	\label{tab:model_architectures}
	\begin{tabular}{|l|c|c|c|c|}
		\hline
		\textbf{Model Name} & \textbf{Params} & \textbf{Base Model} & \textbf{Vision Encoder} & \textbf{Language Model} \\ \hline\hline
		BLIP & 0.25B & VQA-Base & ViT-B/16 & LLAVA (7B) \\ \hline
		BLIP & 0.25B & VQA-Base & ViT-B/16 &  Qwen2-VL(7B) \\ \hline
		InstructBlip & 3B & Flan-T5-XL & ViT-L/14 & LLAVA (7B) \\ \hline
		InstructBlip & 3B & Flan-T5-XL & ViT-L/14 & Qwen2-VL(7B)\\ \hline
		
	\end{tabular}
\end{table*}



\subsection{\textbf{Evaluation Metrics}}
To quantitatively evaluate the performance of the VQA models, we employed two widely used text generation metrics: BLEU and BERTScore. BLEU measures n-gram overlap between the predicted and reference answers, providing a straightforward way to assess exact lexical matches. However, VQA often allows multiple semantically equivalent answers with different wording, which BLEU may not fully capture. To address this, we also used BERTScore, which leverages contextual embeddings from pre-trained language models to evaluate semantic similarity between predictions and references. Together, these metrics provide a complementary assessment: BLEU captures precise lexical alignment, while BERTScore ensures that semantically correct answers are recognized even when phrasing differs. \paragraph{\textbf{BLEU   Score}} The Bilingual Evaluation Understudy (BLEU) score measures n-gram similarity between a machine-generated sentence and reference sentences, penalized by a brevity penalty (BP). Formally, the BLEU score is defined as in Eq.~\ref{eq:bleu}.
\begin{equation}
	\text{BLEU} = \text{BP} \cdot \exp\left(\sum_{n=1}^{N} w_n \log p_n\right)
	\label{eq:bleu}
\end{equation}
where \(p_n\) denotes the modified precision for n-grams of order \(n\), and \(w_n\) represents the corresponding weighting factors, typically set to \(w_n = \frac{1}{N}\).

The brevity penalty (BP) is introduced to penalize hypotheses that are shorter than the reference and is defined as:
\begin{equation}
	\mathrm{BP} =
	\begin{cases}
		1, & \text{if } c > r, \\
		\exp\left(1 - \frac{r}{c}\right), & \text{if } c \le r,
	\end{cases}
\end{equation}
where \(c\) denotes the length of the generated candidate sequence and \(r\) represents the effective reference length.

A higher BLEU score indicates a greater overlap between the generated and reference texts, reflecting better quality of generation. 

\paragraph{\textbf{BERT Score}} BERT Score leverages contextual embeddings from BERT to compare semantic similarity. It computes Precision (P), Recall (R), and an F1-score (F) based on the cosine similarity of tokens. Formally, let \( \mathbf{x} = \{x_i\}_{i=1}^{m} \) and \( \mathbf{y} = \{y_j\}_{j=1}^{n} \) denote the contextualized embeddings of the candidate and reference tokens, respectively. The BERT Score precision and recall are defined as:
\begin{equation}
	P = \frac{1}{m} \sum_{i=1}^{m} \max_{j} \cos(x_i, y_j),
\end{equation}
\begin{equation}
	R = \frac{1}{n} \sum_{j=1}^{n} \max_{i} \cos(y_j, x_i),
\end{equation}
where \( \cos(\cdot, \cdot) \) denotes cosine similarity. The final BERT Score is computed as the harmonic mean of precision and recall:
\begin{equation}
	\mathrm{BERT Score} = \frac{2PR}{P + R}.
\end{equation}

Higher BERT Score values indicate stronger semantic similarity between the generated and reference texts. 

\subsection{\textbf{Quantitative Results}}
To comprehensively evaluate the fine-tuned models, we conducted experiments using two distinct backbone architectures: LLAVA and Qwen2-VL-7B. Both backbones were tested under identical fine-tuning settings to ensure a fair comparison. As shown in Table~\ref{tab:results_comparison_combined}, Qwen2-VL-7B consistently outperformed LLAVA across both BLEU and BERT score metrics. For instance, BLIP fine-tuned with Qwen2-VL-7B achieved a BLEU score of $0.8785$ and a BERTScore of $0.9517$, surpassing its LLAVA counterpart BLEU: $0.8291$, BERT score: $0.9273$. A similar improvement was observed for InstructBLIP, where Qwen2-VL-7B again yielded higher scores, BLEU: $0.6798$, BERT score: $0.7935$, compared to LLAVA BLEU: $0.6345$, BERT score: $0.7612$. These results suggest that Qwen2-VL-7B offers a stronger representational capacity and better semantic alignment for multimodal reasoning tasks, such as WasteVQA.

\begin{table*}[t]
	\centering
	\caption{Performance comparison of fine-tuned models on WasteVQA with LLAVA and Qwen2-VL-7B.}
	\label{tab:results_comparison_combined}
	\begin{tabular}{lcccc}
		\toprule
		\multirow{2}{*}{\textbf{Model}} 
		& \multicolumn{2}{c}{\textbf{LLAVA}} 
		& \multicolumn{2}{c}{\textbf{Qwen2-VL-7B}} \\
		\cmidrule(lr){2-3} \cmidrule(lr){4-5}
		& \textbf{BLEU} & \textbf{BERT} & \textbf{BLEU} & \textbf{BERT} \\
		\midrule
		BLIP         & 0.8291 & 0.9273 & 0.8785 & 0.9517 \\
		InstructBLIP & 0.6345 & 0.7612 & 0.6798 & 0.7935 \\
		\bottomrule
	\end{tabular}
\end{table*}

\begin{figure*}[htbp]
	\centering
	
	\begin{subfigure}{0.30\textwidth}
		\includegraphics[width=\linewidth]{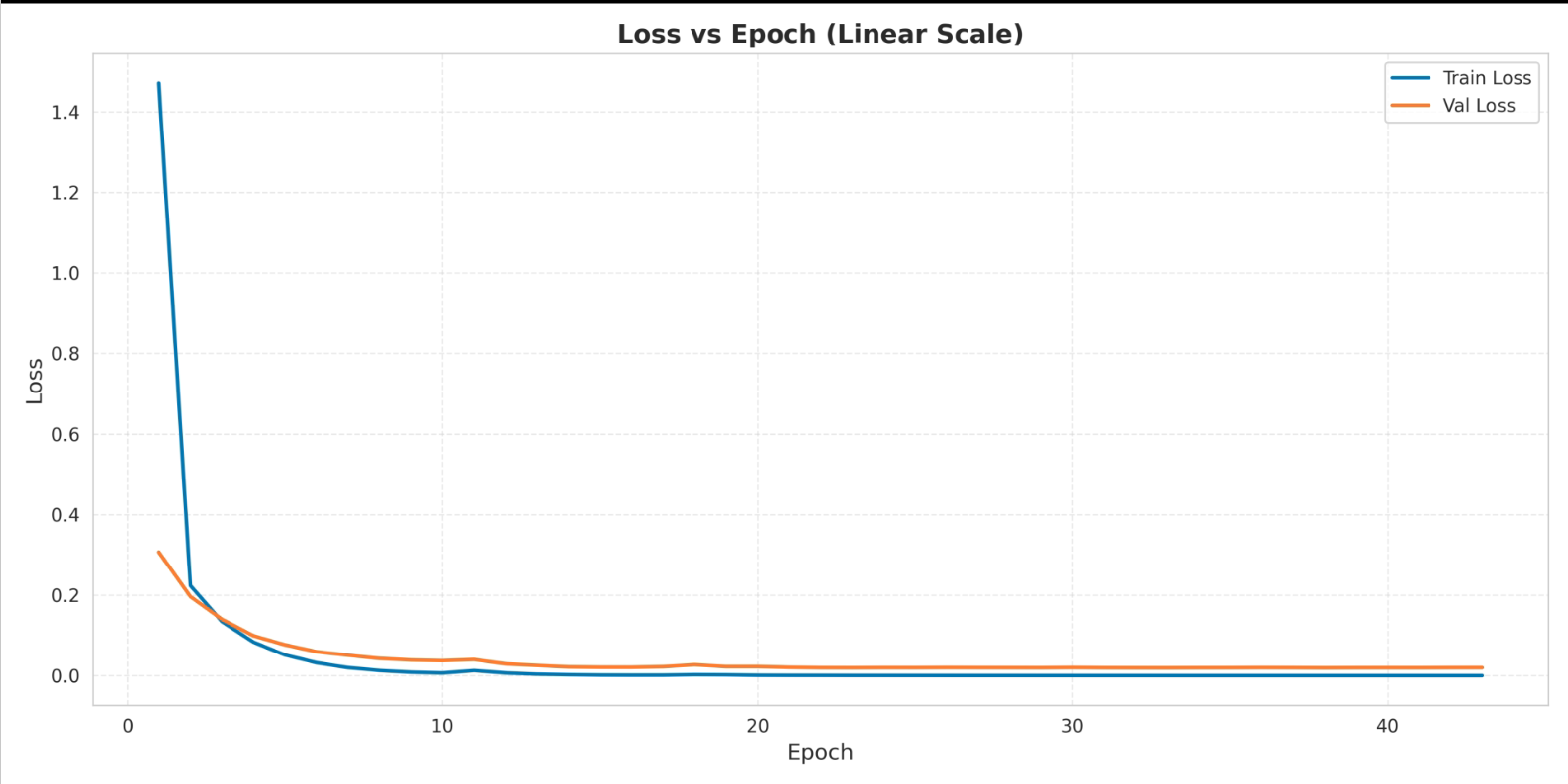}
		\caption{BLIP: Training and validation loss}
		\label{fig:bliploss_linear}
	\end{subfigure}
	\hfill
	\begin{subfigure}{0.30\textwidth}
		\includegraphics[width=\linewidth]{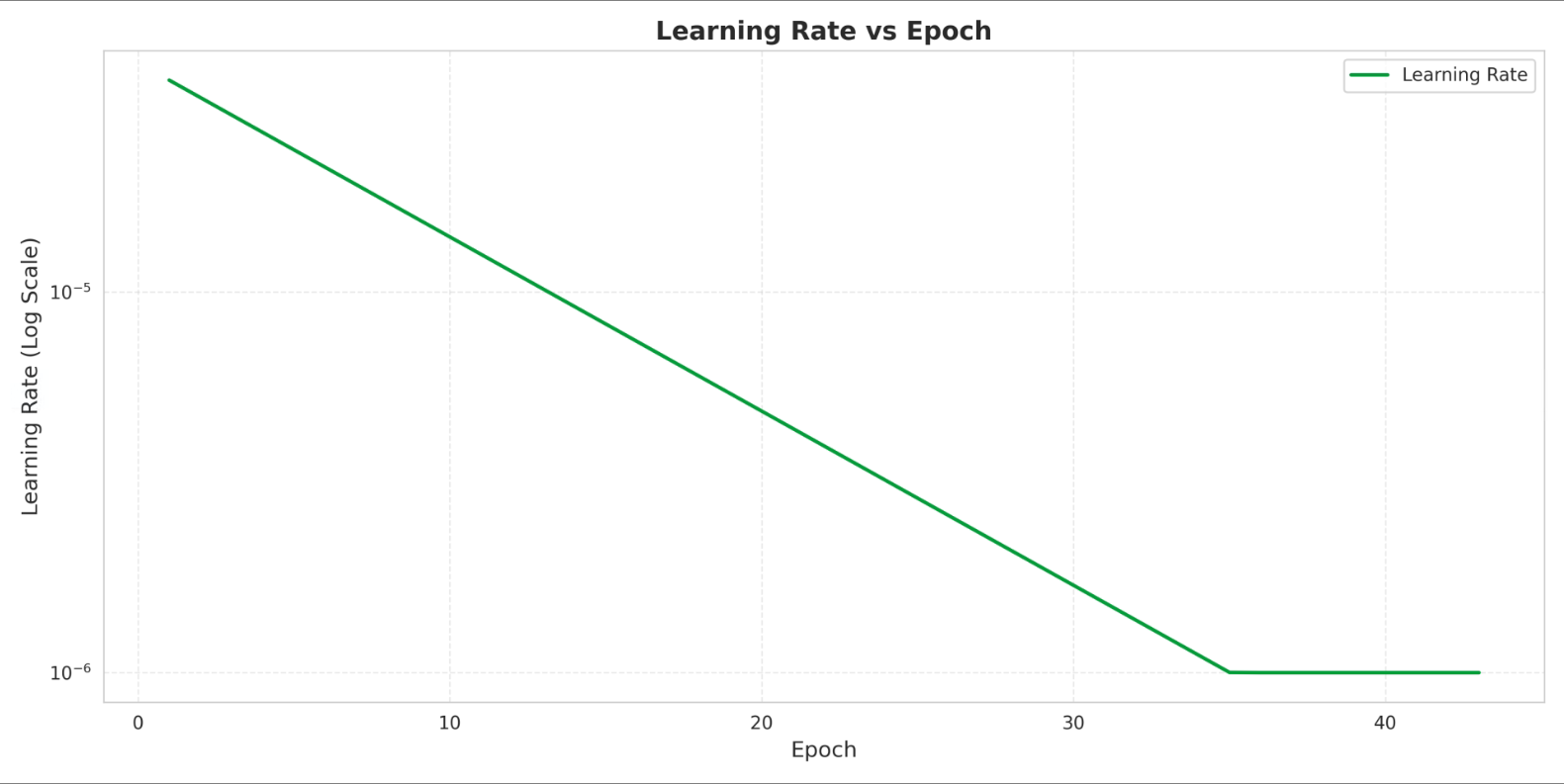}
		\caption{BLIP: Learning rate schedule}
		\label{fig:bliplr_schedule}
	\end{subfigure}
	\hfill
	\begin{subfigure}{0.30\textwidth}
		\includegraphics[width=\linewidth]{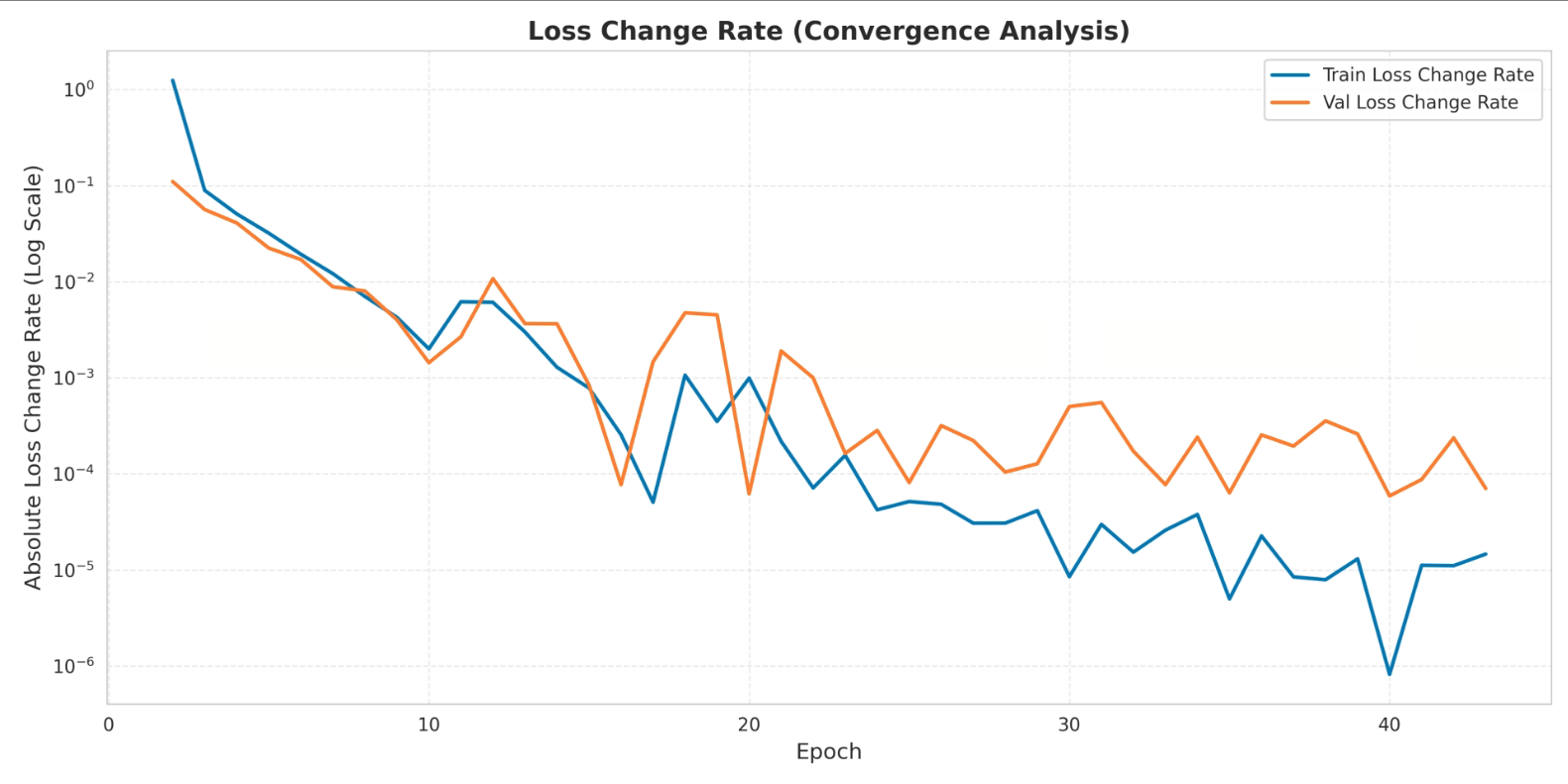}
		\caption{BLIP: Loss change rate}
		\label{fig:bliploss_change}
	\end{subfigure}
	
	\vspace{0.5cm}
	
	\begin{subfigure}{0.30\textwidth}
		\includegraphics[width=\linewidth]{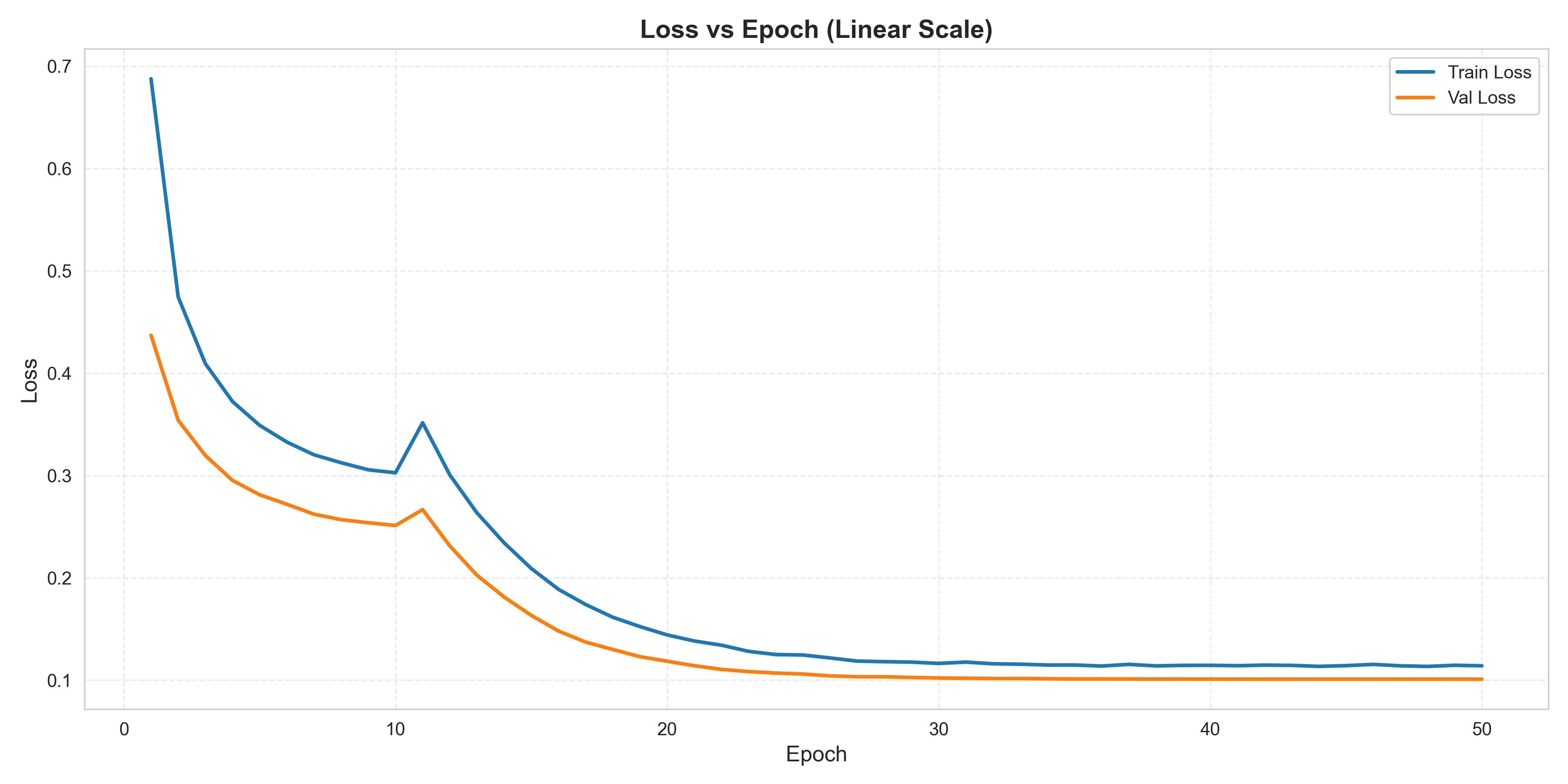}
		\caption{InstructBLIP: Training and validation loss}
		\label{fig:instructbliploss_linear}
	\end{subfigure}
	\hfill
	\begin{subfigure}{0.30\textwidth}
		\includegraphics[width=\linewidth]{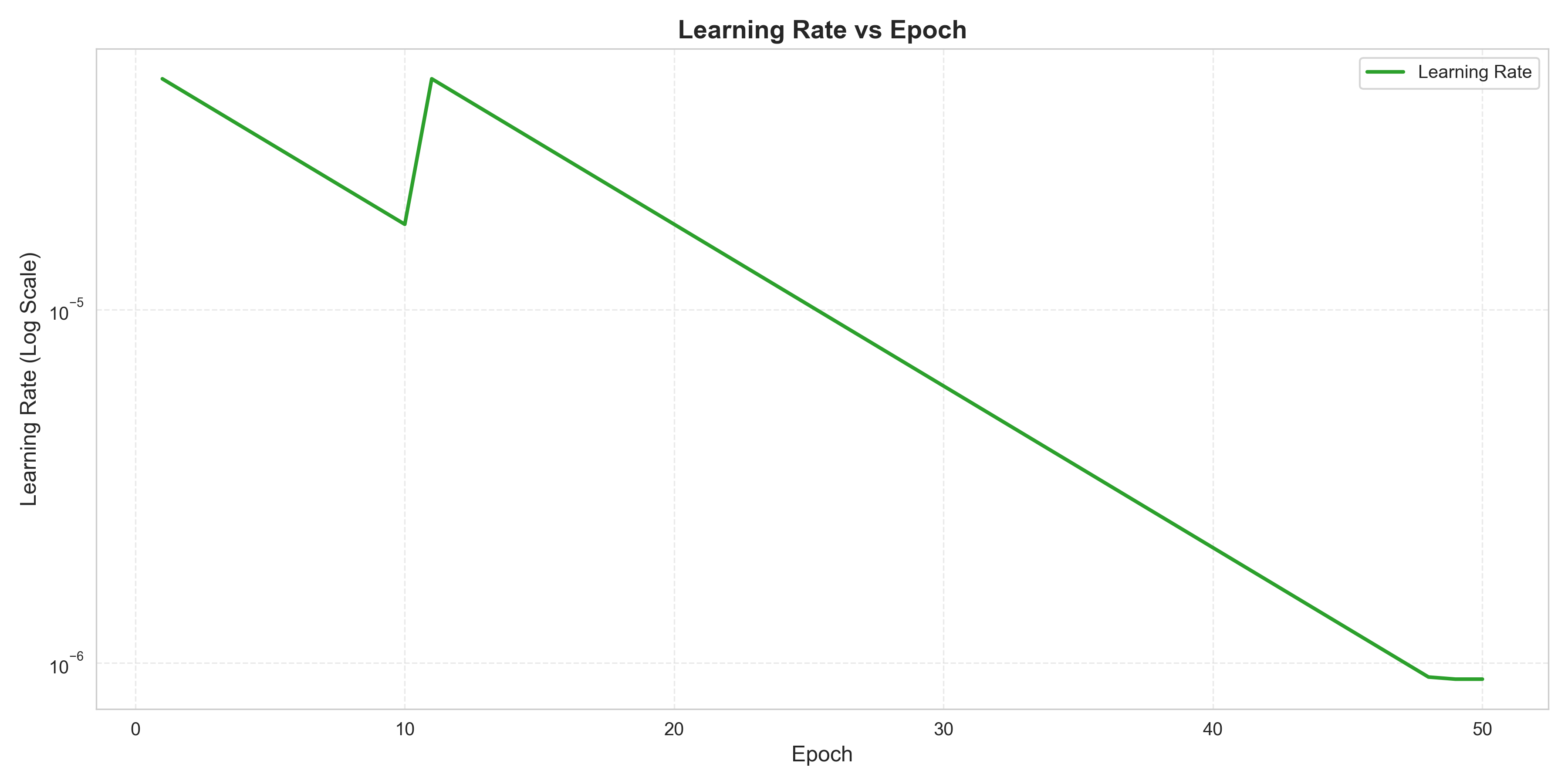}
		\caption{InstructBLIP: Learning rate schedule}
		\label{fig:instructbliplr_schedule}
	\end{subfigure}
	\hfill
	\begin{subfigure}{0.30\textwidth}
		\includegraphics[width=\linewidth]{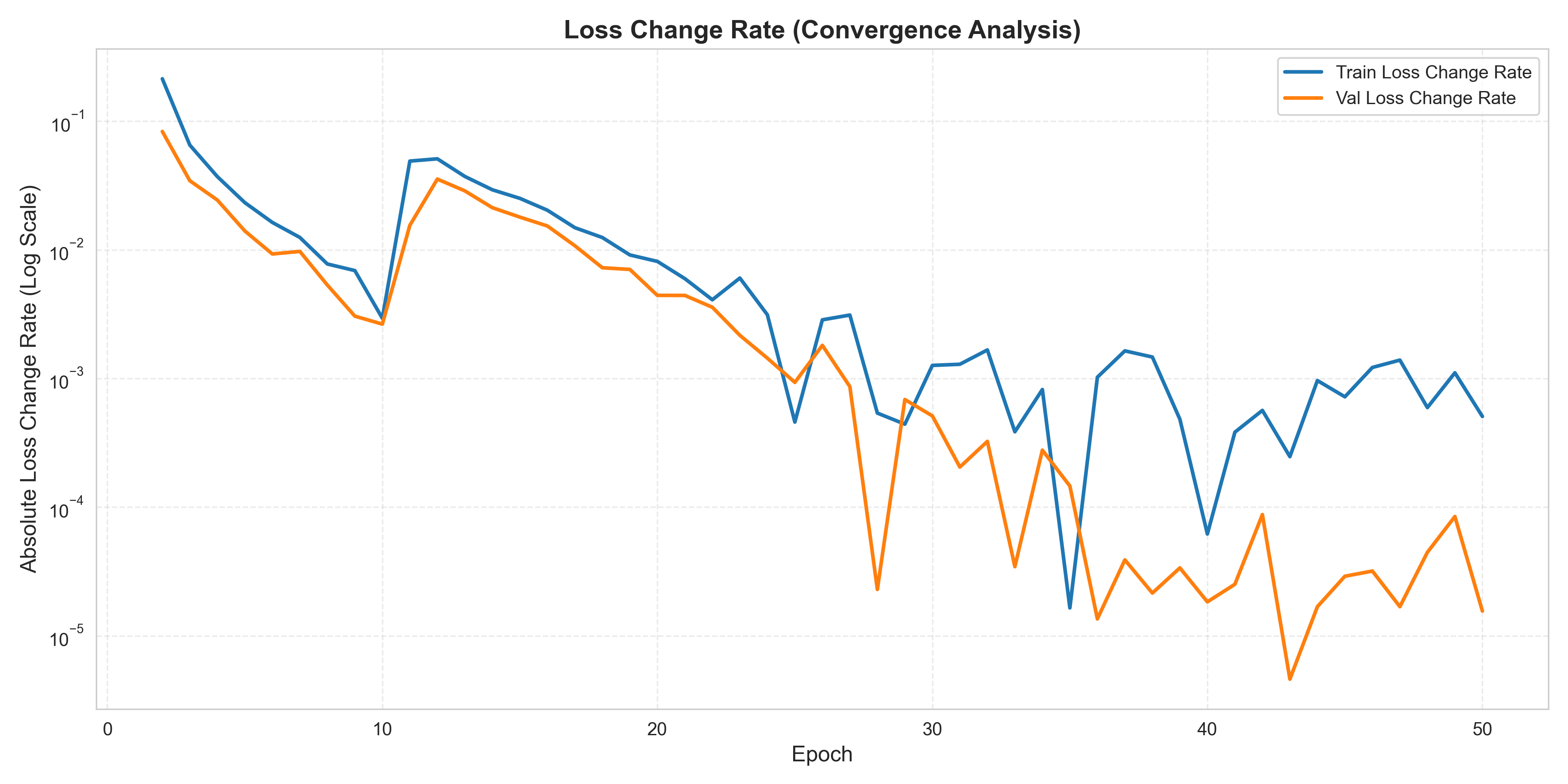}
		\caption{InstructBLIP: Loss change rate}
		\label{fig:instructbliploss_change}
	\end{subfigure}
	
	\caption{Training and validation metrics analysis for the fine-tuned BLIP and InstructBLIP models. The first row presents BLIP performance, while the second row illustrates InstructBLIP performance in terms of training-validation loss, learning rate scheduling, and convergence behavior.}
	\label{fig:blip_instructblip_combined}
\end{figure*}

This observation highlights the importance of selecting a strong vision–language backbone. Qwen2-VL-7B demonstrates superior generalization ability and richer semantic understanding compared to LLaVA, making it more suitable for real-world waste-related visual question answering tasks. To further analyze the optimization behavior, we examine the training and validation loss trajectories of the fine-tuned BLIP model in (Fig.~\ref{fig:blip_instructblip_combined}) first row and the fine-tuned InstructBLIP model (Fig.~\ref{fig:blip_instructblip_combined}) in second row. In both cases, the validation loss closely tracks the training loss throughout the learning process, indicating stable convergence and the absence of overfitting. This close alignment suggests that the learned representations generalize effectively to unseen samples, thereby reinforcing the robustness and reliability of the adopted fine-tuning strategy.



%
%

\subsection{\textbf{Qualitative Results}}
\begin{figure*}[h!]
	\centering
	\includegraphics[width=\textwidth, height=10cm]{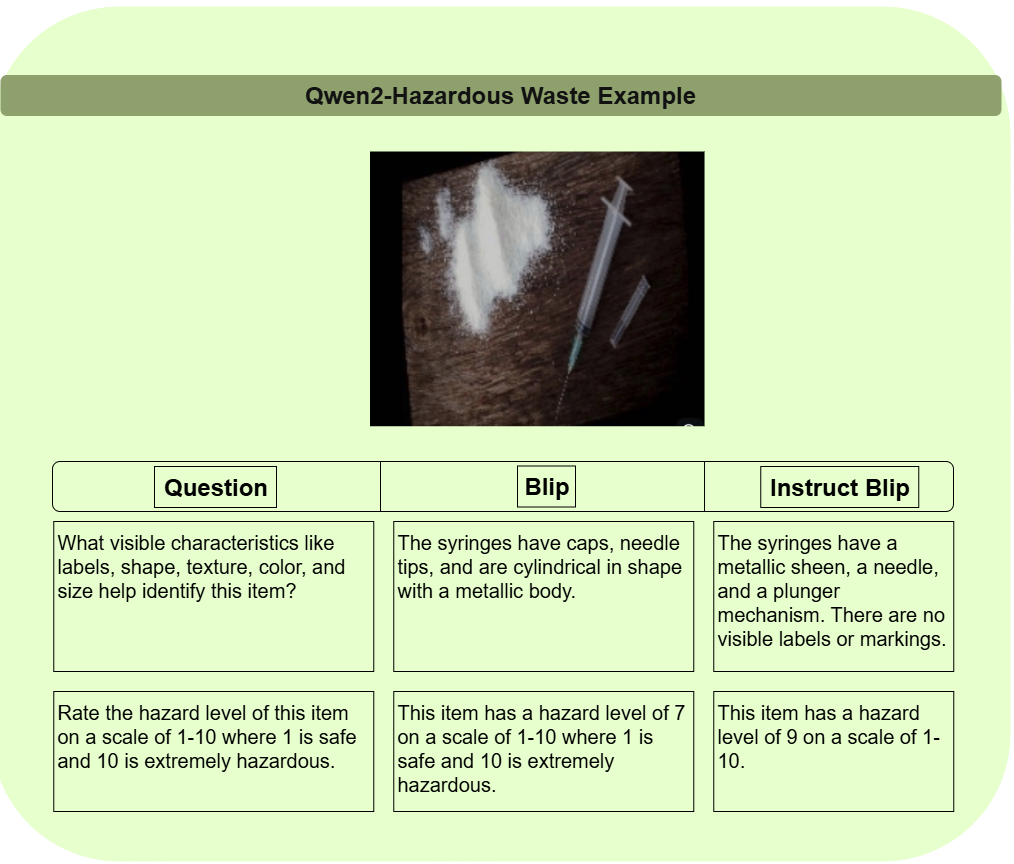}
	\caption{Qualitative comparison of BLIP and InstructBLIP predictions for a hazardous waste example. The figure shows the input image, the question asked, and the distinct responses from each model.}
	\label{fig:qualitative_comparison}
\end{figure*}

In Fig.\ref{fig:qualitative_comparison} illustrates an input image containing a syringe and powder-like substance, along with the corresponding question posed to the models. BLIP provides a surface-level description focused primarily on basic geometric and material attributes (e.g., shape, caps, and metallic body), and assigns a moderate hazard score. In contrast, InstructBLIP produces a more semantically grounded response, explicitly identifying functional components such as the needle, plunger mechanism, and metallic sheen, while also assigning a higher hazard rating. This comparison highlights InstructBLIP’s stronger capabilities in contextual reasoning and risk awareness, which are critical for accurate understanding and assessment of hazardous waste.


\subsection{\textbf{Case Study: Real World Waste Video Analysis}}
To assess the practical applicability of our system in a dynamic setting, we conducted a case study using video input. The video consisted of a sequence of waste items, including plastic wrappers, plastic bags, cardboard pieces, and medical waste such as empty medicine packets. Frames were extracted at a rate of one frame per second (1 FPS), thereby creating a continuous stream of static images that represented real-world waste presentations. For each extracted frame, we employed the LLAVA and Qwen2-VL-7B backbones to generate ground-truth answers for a set of predefined questions, forming a new test set specific to this experiment. Subsequently, the fine-tuned BLIP model, identified as the best-performing model in our main results, was applied to predict answers on these frames.  
The performance of the system on this video-frame dataset was evaluated using three complementary metrics: Precision, F1-Score, and BERT Score. Precision and F1-Score capture classification accuracy and balance between precision and recall, while BERTScore measures semantic similarity in the generated responses. The results, summarized in Table~\ref{tab:case_study_results_combined}, reveal that LLAVA yields slightly higher precision (59.05 vs. 57.20) and F1-Score (54.62 vs. 53.65), whereas Qwen2-VL-7B achieves a better BERTScore (54.84 vs. 50.23), suggesting stronger semantic alignment in its predictions.  
\begin{table*}[t]
	\centering
	\caption{Comparative performance metrics for BLIP and InstructBLIP}
	\label{tab:case_study_results_combined}
	\begin{tabular}{lcccc}
		\toprule
		\multirow{2}{*}{\textbf{Model}} 
		& \multicolumn{2}{c}{\textbf{LLAVA}} 
		& \multicolumn{2}{c}{\textbf{Qwen2-VL-7B}} \\
		\cmidrule(lr){2-3} \cmidrule(lr){4-5}
		& \textbf{BLIP} 
		& \textbf{InstructBLIP} 
		& \textbf{BLIP} 
		& \textbf{InstructBLIP} \\
		\midrule
		Precision   & 59.05 & 63.86 & 57.20 & 60.92 \\
		F1-Score    & 54.62 & 59.28 & 53.65 & 52.25 \\
		BERT Score  & 50.23 & 54.56 & 54.84 & 48.96 \\
		\bottomrule
	\end{tabular}
\end{table*}

Overall, these results indicate that while LLAVA maintains an edge in frame-level precision and balanced classification performance, Qwen2-VL-7B demonstrates a better understanding of semantic aspects related to waste queries. This complementary behavior highlights the importance of model selection depending on whether the application emphasizes strict classification accuracy or deeper semantic reasoning in real-world video-based waste analysis.

\section{Discussion}
The results show a notable performance difference between the fine-tuned BLIP model (BLEU: 0.8291, BERT: 0.9273) and the InstructBlip model (BLEU: 0.6345, BERT: 0.7612). While seemingly counterintuitive given InstructBlip's advanced capabilities, this discrepancy can be attributed to a fundamental mismatch between the model's design and the specific nature of our task.

InstructBLIP is pre-trained for conversational and descriptive instruction-following on a general domain. It excels at generating detailed, human-like prose. Our WasteVQA dataset is a highly specialized, non conversational dataset where the answers are concise, factual, and rule-based. The task is not to describe the image, but to provide a specific, often short, piece of information based on the SWM Rules (e.g., "Green", "No", "Hazard level: 7").

The foundational BLIP model, designed for direct VQA, is better suited for this factual-recall task. Its architecture is more directly optimized for grounding a simple question to an image and producing a short, accurate answer. InstructBLIP's more complex generative process, optimized for dialogue, may produce answers that are syntactically different from the ground truth (e.g., adding conversational fillers) while being semantically similar, which can unfairly penalize it under strict metrics like BLEU.

\section{Conclusion}
Building on this paradigm, our work brings VQA to the environmental domain, focusing specifically on regulatory compliance under India’s Solid Waste Management Rules (2016). We introduce WasteVQA, a regulation-aware VQA dataset comprising 13,500 curated question-answer pairs across 21 waste categories. Unlike standard classification approaches, our method enables models to reason about what an object is and what should be done with it, providing actionable and policy-aligned guidance for disposal. This shift from object recognition to regulation-grounded recommendation addresses a crucial gap in smart waste management systems, demonstrating the real-world applicability of vision-language AI in promoting sustainable practices.

This research successfully demonstrates the potential of integrating deep learning with Visual Question Answering systems to address the challenges of solid waste management. By creating a specialized dataset, "WasteVQA," aligned with India's SWM Rules, we have developed a model capable of providing accurate, context-aware disposal guidance. The superior performance of the standard BLIP model highlights the importance of matching model architecture to task specificity. This work lays the groundwork for practical tools, such as mobile applications, that can empower citizens, improve waste segregation at the source, and contribute to a more sustainable environment. Future work could involve expanding the dataset to include more regional variations and deploying the system in a real-world pilot study.
%
%
\section*{Competing Interests}
The authors declare that they have no competing interests.
\section*{Funding}
This research did not receive any specific grant from funding agencies in the public, commercial, or not-for-profit sectors.

%



%

\end{document}